\documentclass[letterpaper, 10 pt, conference]{ieeeconf}  

\IEEEoverridecommandlockouts                              

\usepackage{multirow}
\usepackage{graphics} 
\usepackage{epsfig} 

\usepackage{graphicx}
\usepackage{color}
\usepackage{url}
\usepackage{algorithm}
\usepackage{algpseudocode}

\title{Derm-T2IM: Harnessing Synthetic Skin Lesion Data via Stable Diffusion Models for Enhanced Skin Disease Classification using ViT and CNN
}

\author{Muhammad Ali Farooq$^{1}$, Wang Yao$^{1}$, Michael Schukat$^{2}$, Mark A Little$^{3}$ and Peter Corcoran$^{1}$
\thanks{*This work is supported by ADAPT - Centre for Digital Content Technology and Enterprise Ireland}
\thanks{$^{1}$M. A. Farooq ({\small muhammadali.farooq@universityofgalway.ie}), W. Yao and P. Corcoran are with College of Science and Engineering, University of Galway, Galway H91 TK33, Ireland}        
\thanks{$^{2}$M. Schukat is with the School of Computer Science, University of Galway, Galway H91 TK33, Ireland}
\thanks{$^{3}$M. A. Little is with Trinity Kidney Centre, Trinity Translational Medicine Institute, Trinity College Dublin, the University of Dublin, St James's Hospital campus, Dublin D08 W9RT, Ireland}
}

\begin{document}

\maketitle
\thispagestyle{empty}
\pagestyle{empty}

\begin{abstract}
This study explores the utilization of Dermatoscopic synthetic data generated through stable diffusion models as a strategy for enhancing the robustness of machine learning model training. Synthetic data generation plays a pivotal role in mitigating challenges associated with limited labeled datasets, thereby facilitating more effective model training. In this context, we aim to incorporate enhanced data transformation techniques by extending the recent success of few-shot learning and a small amount of data representation in text-to-image latent diffusion models. The optimally tuned model is further used for rendering high-quality skin lesion synthetic data with diverse and realistic characteristics, providing a valuable supplement and diversity to the existing training data. We investigate the impact of incorporating newly generated synthetic data into the training pipeline of state-of-art machine learning models, assessing its effectiveness in enhancing model performance and generalization to unseen real-world data. Our experimental results demonstrate the efficacy of the synthetic data generated through stable diffusion models helps in improving the robustness and adaptability of end-to-end CNN and vision transformer models on two different real-world skin lesion datasets. 
\end{abstract}

\begin{clinical}
Synthetic medical imaging datasets play a crucial role in advancing healthcare by supporting education, research, algorithm development, and addressing privacy concerns. They contribute to the improvement of medical imaging technologies and the development of more accurate and robust diagnostic tools.
\end{clinical}
\section{INTRODUCTION} \label{introduction}
Fatal skin disease, especially malignant skin cancer has emerged as a significant public health concern and is steadily gaining serious medical attention in recent years. The rising incidence and recent statistics of skin cancer~\cite{SkinCancerStatistics} can be attributed to various factors, including increased ultraviolet (UV) radiation exposure due to changing environmental patterns and lifestyle habits. Sun exposure, indoor tanning, and inadequate protection against UV rays significantly contribute to the proliferation of skin cancer cases. Furthermore, an aging population, genetic predispositions, and a lack of awareness regarding skin cancer risk factors further exacerbate the issue. Skin cancer not only poses a substantial threat to individual health and well-being but also exerts a considerable economic burden on healthcare systems worldwide. Therefore, it is imperative to emphasize outreach efforts, promote sun-safe behaviours, and encourage regular skin examinations to mitigate the impact of this growing health problem. Early detection, timely intervention, and proactive preventive measures are essential in reducing the incidence and mortality rates associated with skin cancer. Lots of recent studies have shown shocking statistics~\cite{Cancerstatistics2023,Burden,rashid2022skin} related to the increased number of skin cancer cases throughout the world. 

Large scale medical imaging datasets play a vital role in development and deployment of efficient AI based computer aided diagnosis systems which can be effectively used by doctors and professionals for prompt diagnosis of various skin diseases~\cite{pathania2022non} including malignant skin cancer~\cite{ricotti2009malignant}. Acquiring real-world skin lesion imaging data for detecting various skin disorders is indeed valuable for research and advancements in healthcare. However, acquiring such types of datasets is often challenging and further utilizing these datasets for research and analysis comes with several limitations as discussed in Section~\ref{background} that can impact the accuracy, diversity, and applicability in various contexts.

Keeping this in view, in this research work we have focused on utilizing AI-based generative stable diffusion models~\cite{carlini2023extracting} for generating large-scale synthetic skin lesion data by unveiling the power of large text-to-image diffusion models~\cite{saharia2022photorealistic}. This novel framework is referred to as Dermatoscopic Text-to-Image Model (Derm-T2IM) that uses natural language text prompts as an input and produces high quality malignant (including melanoma, basal cell carcinoma and squamous cell carcinoma) and benign (including nevus) lesion imaging data as the output. This framework can effectively comprehend the requirements of real-world large-scale skin imaging datasets. Incorporating image synthesis techniques into medical research can enhance data utilization, improve machine learning model performance, and accelerate advancements in the diagnosis, treatment, and understanding of various medical conditions. Further the data generated using diffusion based generative algorithms works as a reliable alternative to real world medical imaging data by providing various advantages such that synthetic images can be created to represent a wide range of variations in anatomy, pathology, dermoscopy, medical imaging modalities, and clinical scenarios. Secondly image synthesis allows for the creation of de-identified and privacy-preserving synthetic medical data, enabling secure sharing of data for research and collaboration without risking patient privacy or violating regulatory requirements. Further synthetic images can serve as a benchmark for evaluating the performance of various image processing, reconstruction, and enhancement algorithms, enabling a fair comparison of different methodologies in a controlled setting. Last but not least generating synthetic data is often less expensive and time-consuming compared to collecting real patient data. This cost-effectiveness can accelerate research and development in the field of medical imaging. 

In this work we have used a few shot data learning technique~\cite{wang2020generalizing} via pretrained stable diffusion models~\cite{rombach2021highresolution} to address the challenges of learning patterns from a controlled amount of dermatoscopic data. The tuned model is then used for generating high-quality synthetic skin lesion data samples. The advantages of using few-shot adaptation of stable diffusion models include the ability to generate diverse and realistic data samples, which can lead to better generalization and performance on tasks with limited training data. To accomplish this task, we have employed DreamBooth~\cite{ruiz2023dreambooth} for tuning a customized text-to-image diffusion model capable of generating high quality skin lesion data. Section 3 details the proposed algorithm and complete working methodology proposed in this research work. The quality of generated synthetic data is cross-validated by fine-tuning two state-of-the-art deep learning classifiers which include vision transformers~\cite{dosovitskiy2020image} and end-to-end pretrained CNN MobileNet-v2~\cite{howard2017mobilenets} architecture. The complete experimental results are detailed in Section~\ref{experiment}. The core contributions of this research work are listed below.

\begin{itemize}
    \item Adaptation and optimal tuning of stable diffusion models with as low as 1400 images per class for developing a customized dermatoscopic text to image (T2I) diffusion models for rendering large scale synthetic dermatoscopic data based on user text prompts.
    \item The tuned T2I model is open-sourced via hugging face library [Link: https://huggingface.co/MAli-Farooq/Derm-T2IM] for research community and medical professionals to render more synthetic dermatoscopic skin lesion data as per their requirements. 
    \item A novel large scale synthetic skin lesion data with various smart transformations which includes skin lesions of different classes, manipulating the size of moles, and rendering multiple moles, is open-sourced. The overall data comprises 3k malignant skin mole samples and 3k benign moles.
    \item Further validating the efficacy of newly generated skin lesion data by fine tuning vision transformers and end-to-end pre-trained CNN MobileNet V2~\cite{howard2017mobilenets} with different settings. The performance of tuned classifiers is cross validated against ISIS Archive and HAM10000 skin cancer dataset with the highest overall accuracy of 93.03\% using vision transformers. 
\end{itemize}

\section{Background}\label{background}

Skin lesion detection and classification are crucial in dermatology for the early diagnosis of various skin diseases, including fatal skin cancers such as melanoma~\cite{jones2022artificial,farooq2016automatic,bhatt2023state}. Deep Convolutional Neural Networks (CNNs), such as EfficientNet, DenseNet, ResNet, and Inception, and MobileNet V2 have been widely used for robust skin lesion feature extraction and classification~\cite{gairola2022exploring}. However deep learning models are often considered data-hungry because they typically require a large amount of training data to perform effectively. Table~\ref{tab1} provides a list of large-scale publicly available skin imaging datasets. However, real-world skin cancer datasets used in research and clinical applications can present several challenges and limitations. These problems can affect the quality, representativeness, and applicability of the data. Some common problems with large-scale real-world skin lesion include: 

\begin{table*}[!tbp]
\centering
\caption{Datasets}
\label{tab1}
\resizebox{\linewidth}{!}{
\begin{tabular}{|l|l|l|l|}
\hline
S.No &
  Dataset Name &
  Dataset Details &
  Clinical Diagnostic Attributes \\ \hline
1 &
  \begin{tabular}[c]{@{}l@{}}ISIC \\ (International\\ Skin Imaging \\ Collaboration)\\ Archive~\cite{Melanoma} \end{tabular} &
  \begin{tabular}[c]{@{}l@{}}The ISIC Archive is one of the \\ most comprehensive resources \\ for dermatology-related image \\ datasets. \\ It includes a variety of skin \\ lesion images, with labels for\\ conditions like melanoma, nevus,\\ and basal cell carcinoma.\end{tabular} &
  \begin{tabular}[c]{@{}l@{}}nevus, melanoma, basal cell carcinoma, \\ seborrheic keratosis, squamous cell \\ carcinoma, actinic keratosis, pigmented\\ benign keratosis, solar lentigo,\\ dermatofibroma, vascular lesion, \\ lichenoid keratosis, acrochordon, \\ lentigo NOS, atypical melanocytic \\ proliferation, AIMP, verruca, angioma, \\ lentigo simplex, melanoma metastasis, \\ other (33)\end{tabular} \\ \hline
2 &
  PH$^{2}$ Dataset~\cite{mendoncca2013ph} &
  \begin{tabular}[c]{@{}l@{}}The PH$^{2}$ Dataset consists of \\ dermoscopic images of common \\ pigmented skin lesions, along \\ with ground truth data. \\ It is commonly used for research \\ in melanoma detection.\end{tabular} &
  \begin{tabular}[c]{@{}l@{}}Common Nevus, atypical nevus, \\ melanoma\end{tabular} \\ \hline
3 &
  \begin{tabular}[c]{@{}l@{}}HAM10000 \\ (Human \\ Against \\ Machine \\ with 10,000 \\ training \\ images) \\ Dataset~\cite{tschandl2018ham10000}\end{tabular} &
  \begin{tabular}[c]{@{}l@{}}This dataset contains over 10,000 \\ high-quality dermatoscopic images \\ with annotations for different \\ skin lesion categories, including \\ melanoma, and nevus.\\ The overall dataset includes a \\ representative collection of a \\ wide range of important diagnostic \\ categories in the realm of \\ pigmented lesions\end{tabular} &
  \begin{tabular}[c]{@{}l@{}}Ctinic keratosis and intraepithelial \\ carcinoma, Bowen's disease, basal cell \\ carcinoma, benign keratosis-like lesions\\ (solar lentigines / seborrheic keratoses\\ and lichen-planus like keratosis, \\ dermatofibroma, melanoma, melanocytic \\ nevi and vascular lesions (angiomas, \\ angiokeratomas, pyogenic granulomas \\ and hemorrhage\end{tabular} \\ \hline
4 &
  \begin{tabular}[c]{@{}l@{}}Dermofit \\ Image \\ Library~\cite{inbook}\end{tabular} &
  \begin{tabular}[c]{@{}l@{}}The dataset provides high quality \\ skin lesion images for research \\ purposes in computer science and \\ medical imaging.\\ The Dermofit Image Library is a \\ collection of 1,300 focal high \\ quality skin lesion images \\ collected under standardised \\ conditions with internal colour \\ standards.\end{tabular} &
  \begin{tabular}[c]{@{}l@{}}Ctinic Keratosis, Basal Cell Carcinoma,\\ Melanocytic Nevus (mole), Seborrhoeic \\ Keratosis, Squamous Cell Carcinoma, \\ Intraepithelial Carcinoma, Pyogenic \\ Granuloma, Haemangioma \\ Dermatofibroma, Malignant Melanoma\end{tabular} \\ \hline
5 &
  \begin{tabular}[c]{@{}l@{}}DermIS\\ (Dermatology \\ Information \\ System) \\ Dataset~\cite{Dermatology}\end{tabular} &
  \begin{tabular}[c]{@{}l@{}}It is one of largest dermatology \\ datasets available online. \\ DermIS offers a collection of \\ clinical images and cases related \\ to dermatology.\end{tabular} &
  \begin{tabular}[c]{@{}l@{}}Acrolentiginous Melanoma, Basal Cell \\ Carcinoma, Hematoma, Melanocytic Nevus\\ (mole), and others\end{tabular} \\ \hline
\end{tabular}%
}
\end{table*}

\begin{itemize}
    \item Data Quality and Variability: Skin cancer datasets often suffer from varying image quality, lighting conditions, resolutions, and perspectives. Inconsistent image quality can affect the accuracy and reliability of machine learning models.
    \item Data Imbalance: Dermatoscopic datasets may exhibit imbalanced class distributions, with some types of skin lesions being more prevalent than others. This can bias machine learning models and affect their ability to accurately classify rare skin conditions.
    \item Inter-Observer Variability: Different medical professionals might interpret and label skin lesions differently. This can lead to discrepancies in the dataset, affecting the ground truth and potentially influencing the performance of machine learning algorithms.
    \item Heterogeneity and Standardization: Skin cancer datasets often come from diverse sources, resulting in heterogeneous data in terms of image formats, annotations, and metadata. Standardizing the data is challenging, but critical for effective analysis and model training.
    \item Lack of Annotated Data: Annotating skin lesion images requires expert dermatologists, and obtaining a large-annotated dataset can be time-consuming and expensive. Limited availability of annotated data hinders the training of accurate machine learning models.
    \item Privacy and Ethical Concerns: Medical datasets, including pigmented skin lesions, contain sensitive information about individuals. Ensuring privacy and complying with ethical guidelines while sharing or using such data poses a significant challenge. Further in the European Region (EU) General Data Protection Regulation (GDPR) sets stringent rules regarding the collection, processing, and storage of sensitive medical data.
    \item Longitudinal Data: Tracking the progression of skin lesions over time is essential for a comprehensive understanding of skin cancer. However, obtaining longitudinal data with consistent patient follow-ups can be difficult and may limit research insights.
    \item Diversity of Skin Types and Ethnicities: Skin cancer affects individuals with various skin types and ethnic backgrounds differently. The lack of diversity in the dataset may result in biased models that do not perform well for certain populations.
    \item Limited Metadata and Clinical Information: Many skin cancer datasets lack detailed clinical information such as patient demographics, medical history, and other relevant factors. This limits the ability to conduct comprehensive analyses and develop more context-aware models.
    \item Generalization to Real-World Settings: Machine learning models, especially deep learning networks trained on controlled datasets may not always generalize well to real-world clinical settings. The models need to be robust and applicable to diverse clinical environments for effective diagnosis and decision support.
\end{itemize}

Addressing these challenges requires collaboration among clinicians, data scientists, and researchers to improve data quality, standardization, and diversity in skin lesion datasets which is often very laborious and time-consuming process.

\subsection{Data Augmentation in Medical Imaging using Computer Vision Algorithms}

To address the challenges discussed in Section~\ref{background}, we have concentrated on working with digital image synthesis for dermatoscopy. It involves generating synthetic data, with precise details of skin lesions. Further, dermatoscopic image synthesis serves various purposes, including augmenting datasets, simulating skin conditions, and enhancing the training of machine learning models for accurate skin lesion analysis. This can be attained by using various advanced image generation techniques, including generative adversarial networks (GANs)~\cite{skandarani2023gans}, variational autoencoders (VAEs)~\cite{pesteie2019adaptive}, physics-based simulations and recently emerged text-to-image stable diffusion models~\cite{kazerouni2023diffusion}. We can find several studies where the research community have particularly focused on using GAN’s for medical image synthesis~\cite{shin2018medical,nie2018medical} however GANs are not inherently invertible. Unlike some generative models, such as VAEs, GANs do not provide a straightforward way to map generated data back to the latent space. This can be a limitation in critical applications such as medical data. 

Synthetic image generation through a stable diffusion model can augment medical imaging datasets. This is especially valuable when there's a scarcity of data for specific conditions and when explicit diversity is required in existing medical imaging modalities. We can find published studies where the research community have proposed digital image synthesis for various medical imaging applications~\cite{akrout2023diffusion,dalmaz2022resvit,liang2022sketch,li2022neural}. Akrout et al.~\cite{akrout2023diffusion} proposed text to image synthesis method which is closest to our work which demonstrates the effectiveness of generative models, specifically probabilistic diffusion models (DPMs), in generating high-quality synthetic images of macroscopic skin diseases. This is achieved by conditioning the generation process on text prompt inputs, allowing for fine-grained control of the image generation process. However, there approach depends on utilizing large number of data samples for robust fine-tuning of pretrained stable diffusion models. Further authors do not discuss much about the computational efficiency of the proposed model. Moreover, at this stage the authors have not open-sourced the synthetic macroscopic skin diseases dataset and tuned models. In another study~\cite{de2023medical}, authors demonstrated that pre-trained Stable Diffusion models, originally trained on natural images, can be adapted for various medical imaging modalities by training text embeddings with textual inversion. The authors have validated the approach on three different medical imaging modalities which include prostate MRI images, chest x-ray and lymph node sections histopathology images. Further experimental analysis shows that synthetic data helps by achieving a 2\% increase in diagnostic accuracy (AUC) for detecting prostate cancer on MRI, from 0.78 to 0.80. Shavlokhova et al.~\cite{shavlokhova2023finetuning} fine-tuned the GLIDE stable diffusion model on different clinical diagnostic entities, including melanoma and benign melanocytic nevi. Authors used 10,015 dermoscopic images for this purpose thus relying on sufficient seed training data. They created photorealistic synthetic samples of each diagnostic entity using tuned models. For evaluation purposes authors have used various image quality metrics which includes SSIM, PSNR, and FID scores.

Our work extends on this by exploring few-shot medical image generation, by training with limited yet high-quality data samples. Additionally, we demonstrate the efficacy of the proposed approach by indulging various smart transformations using text prompts thus reflecting the effectiveness of tuned text-to-image diffusion model. In contrast to most other studies, we intentionally do not train from scratch and use small datasets to explore the feasibility of fine-tuning diffusion in low-data and low-compute environments.

\section{Building Derm-T2IM}
\begin{figure*}[thpb]
    \centering
    \includegraphics[width=0.9\linewidth]{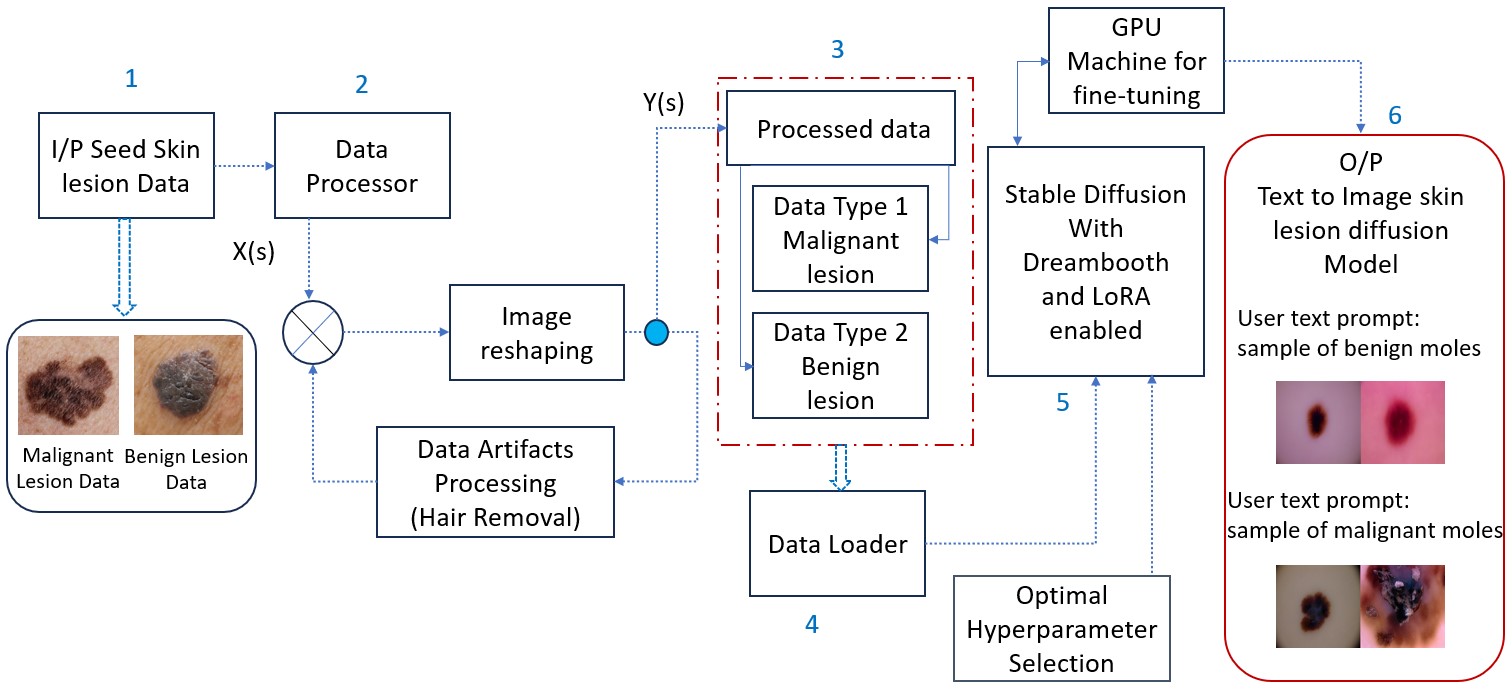}
    \caption{Comprehensive block diagram representation of proposed methodology.}
    \label{framework}
\end{figure*}

This section will describe the proposed methodology adapted in this work for generating large scale high quality synthetic skin lesion data. As mentioned in Section~\ref{introduction} we have primarily focused on using small amount of skin lesion representation and further few shot~\cite{ruiz2023dreambooth} learning method for fine-tuning text-to-image stable diffusion models. Probabilistic Diffusion models~\cite{ho2020denoising} are a class of AI based generative models used to model complex probability distributions. They capture the dynamics of data by iteratively diffusing a data point such as an image towards the data distribution through a series of transformations. These models are particularly useful for generating realistic and high-quality data samples. Stable diffusion models based on a particular type of diffusion model called Latent Diffusion model~\cite{rombach2022high}, are a variant of diffusion models designed to provide more stability and improved training characteristics. They aim to address issues like training instability and mode collapse often encountered with standard diffusion models. Fig.~\ref{framework} shows the block diagram representation of the proposed methodology.

\begin{figure}[thpb]
    \centering
    \includegraphics[width=0.9\linewidth]{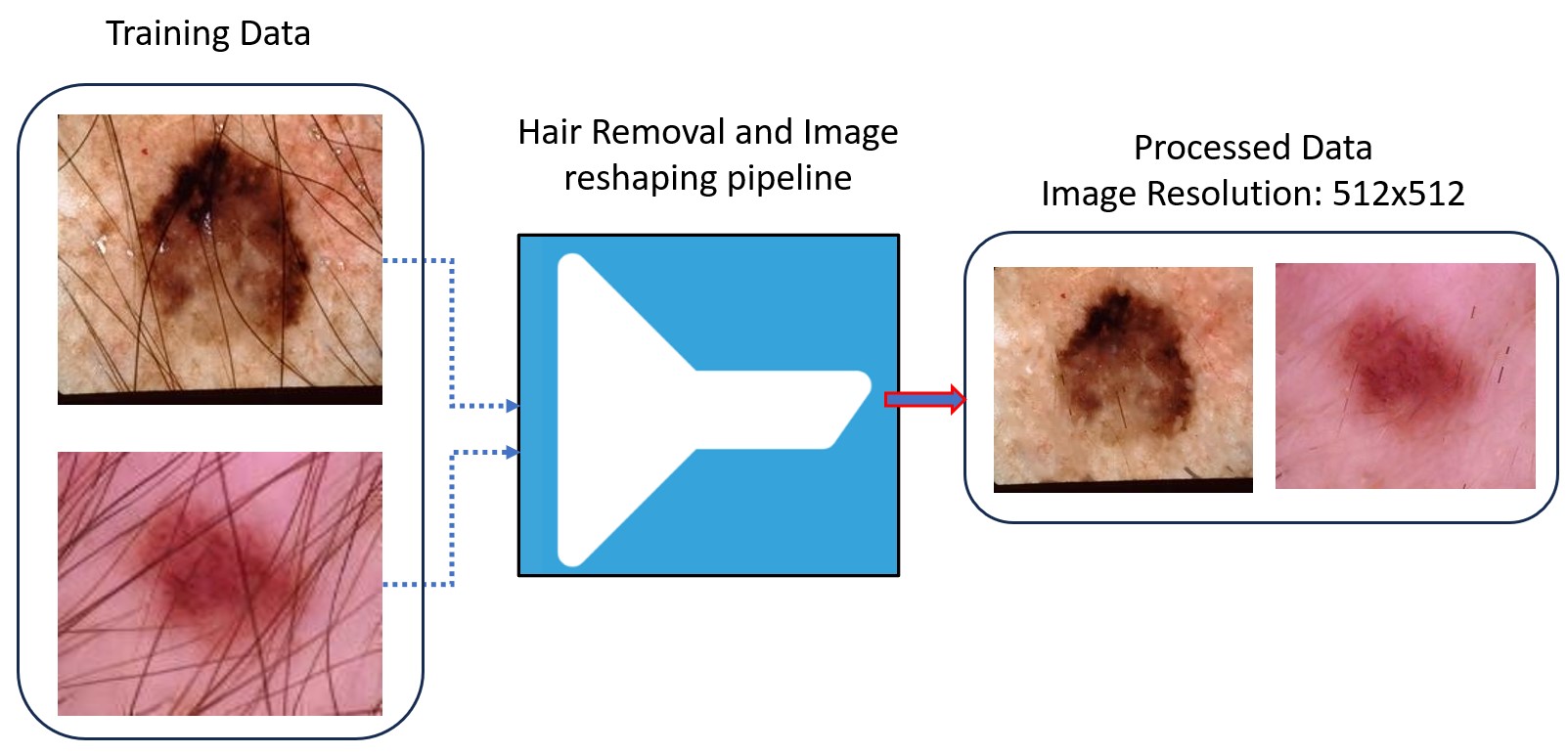}
    \caption{Skin mole hair removal data processing outputs on two different cases generated using dull razor software.}
    \label{fig_2}
\end{figure}

As it can be seen from Fig.~\ref{framework} the first step is preparing the input data (seed data) for training the models. In this work we have used International Skin Imaging Collaboration (ISIC) Archive data~\cite{Melanoma}. The ISIC Archive was created to establish an extensive public repository of meticulously annotated high-quality skin images. This resource provides clinicians and educators in enhancing diagnostic capabilities and delivering clinical assistance for the detection of skin cancer. The selected set data is divided in binary classes which includes malignant and benign mole data. A total of 2800 data samples with 1400 benign skin mole images and 1400 malignant skin moles were used for this purpose. The second step includes data preprocessing thus preparing the data for feeding it to the data loader. Data preprocessing incorporates resizing the image size to $512 \times 512$ as input tensor shape and further removing data clutters such that unwanted skin hairs covering the lesion region of interest thus properly focusing on lesion area. To accomplish this, we have used open-source dull razor software~\cite{lee1997dullrazor}. Fig.~\ref{fig_2} shows the hair removal processing results on two sample cases using dull razor application. 

The third and fourth step include using the processed data and passing it to the data loader. The processed data is divided in two different data loading concepts which includes malignant skin lesion data and benign skin lesion data. 

\subsection{Stable Diffusion Training Components}
As mentioned in Section~\ref{introduction} we have incorporated Dreambooth tool which is based on few shot data learning methods for fine-tuning text-to-image stable diffusion models (T2IM). The core idea is to condition the image generation process on a given image text description. Equation 1 represents the simplified conceptual equation to illustrate the T2IM. Further Low-Rank Adaptation (LoRA) is enabled for performing stable training process. The details of these tools are provided below.
$$
I = G(T, \theta) + \varepsilon \eqno{(1)}
$$

Where $G$ is the generator function that takes the textual description $T$ and model parameters $\theta$ to produce an image $I$. $\varepsilon$ represents random sampling noise added during the image generation process.

\noindent \textbf{Dreambooth}

Dreambooth, which is being developed by Google research and Boston University researchers offers the unique advantage of utilizing a small set of seed training data depicting a particular data class/subject, by adapting a pre-trained text-to-image Imagen~\cite{saharia2022photorealistic} model to associate a distinct identifier with that specific subject. After integrating the subject into the model's output space, this identifier becomes a tool for generating new images of the subject within various contextual settings. The approach leverages the inherent semantic knowledge within the model, along with a novel self-generated class-specific prior preservation loss, thus allowing to create representations of the subject in a wide range of scenes, poses, perspectives, and lighting conditions that were not present in the original reference images. In this work we have adapted the same tool for generating photo-realistic skin lesion data thus validating its effectiveness in medical imaging applications.

\noindent \textbf{LoRA}

 \begin{algorithm*}[ht]
\caption{Derm-T2IM Training Algorithm}
\label{pesudo}
\begin{algorithmic}[1]
    \Require
        Text descriptions $T$;
        Few-shot image data $x(t)$ (Binary Classes: 0, 1).

    \State Preprocess the text descriptions and few-shot image data $x(t)$;
    \State Initialize the model parameters $(\theta)$;
    \For {$i = 1$ to $n$}
        \State Encode the text descriptions $T$ using a text encoder $\iota$;
        \State Conditioning vector $C = \iota(P)$ generated using a text encoder $\iota$ and a text prompt $P$;
        \State Encode the few-shot image data using an image encoder $\varepsilon$;
        \State Combine the encoded text $\iota$ and image features;
        \State Define the noise Scheduler $\delta_t$;
        \State Generate initial images using a generator network $G(T,\theta)$;
        \State Compute the loss between the generated images and the real images;
        \State Update the generator network $G(T,\theta)$ using backpropagation and gradient descent.
    \EndFor
    \Ensure
        $I=G(T,\theta)+\varepsilon$, Generated images from new text descriptions T and random sampling noise.    
\end{algorithmic}
\end{algorithm*}

Utilizing Low-Rank Adaptation~\cite{hu2022lora} in stable diffusion during the training phase is a valuable tool for improving the performance and stability of generative models. Equation 2 shows the incorporating LoRA while fine-tuning stable diffusion models. $\beta$ is the merging ratio. It works by scaling beta from 0 to 1. Setting beta to 0 is the same as using the original model and setting beta to 1 is the same as using the fully fine-tuned model. It allows the efficient incorporation of new data while leveraging the pretrained knowledge in the model's parameters. LoRA enhances training efficiency and significantly reduces the computational hardware requirements, up to threefold, particularly when employing adaptive optimizers. This is achieved by eliminating the need to compute gradients and manage optimizer states for the majority of parameters. As an alternative the focus is on optimizing the compact, injected low-rank matrices, resulting in streamlined and resource-efficient training process. This approach is especially significant in scenarios where retraining from scratch is impractical and resource intensive. In this work we have used Low-rank adaptation for optimal fine-tuning of pretrained stable diffusion models with a lower-dimensional representation based on new data type, thus without fully retraining the entire model. 

$$
W(x) = W_{0} + \beta\Delta W \eqno{(2)}
$$

\noindent \textbf{Tuning Derm-T21M}

Our primary focus is building customized diffusion model using a pre-trained text-to-image diffusion model, which takes a few-shot image data $x(t)$, along with a conditioning vector C generated through a text encoder based on a given text prompt P. This model, when provided with these inputs, and further trained using the mentioned training components, produces fully synthetic images. The below pseudocode shows the complete training description for building Derm-T2IM.

\subsection{Training Parameter Selection}
The next step includes selecting an optimal set of training parameters. Initially we conducted a lot of experiments to analyse the effect of different settings in Dreambooth since it plays a critical role in achieving successful and effective training results. Although the training process relies on numerous training parameters, in this section, we have discussed the most important parameters which are detailed in Table~\ref{tab1} that were shortlisted during the fine-tuning phase of stable diffusion using Dreambooth tool.

\begin{table}
\begin{center}
\caption{Hyper-parameter selection}
\label{tab2}
\resizebox{\linewidth}{!}{
\begin{tabular}{|c|c|}
\hline 
Training Parameter                   & Details             \\ 
\hline 
Training Batch Size     & 2             \\ \hline
Training Epochs & 120             \\ \hline
Gradient accumulation steps                  & 1               \\ \hline
Optimizer         & 8-bit, AdamW              \\ \hline
Minimum learning rate     &  $1e^{-6}$              \\ \hline
Learning rate & $2e^{-6}$ \\ \hline
LoRA learning rate & $1e^{-4}$ \\ \hline
Noise Scheduler                     & \begin{tabular}[c]{@{}c@{}}Discrete Denoising\\  Scheduler(DDS)\end{tabular} \\ \hline 
Mixed precision selection             & FP16              \\ \hline
Learning Scheduler             & \begin{tabular}[c]{@{}c@{}}Constant with\\ warmup\end{tabular} \\ \hline 
Model input size (Resolution)                   & 512              \\  
\hline 
\end{tabular}
}
\end{center}
\end{table}
\begin{figure}[thpb]
    \centering
    \includegraphics[width=0.8\linewidth]{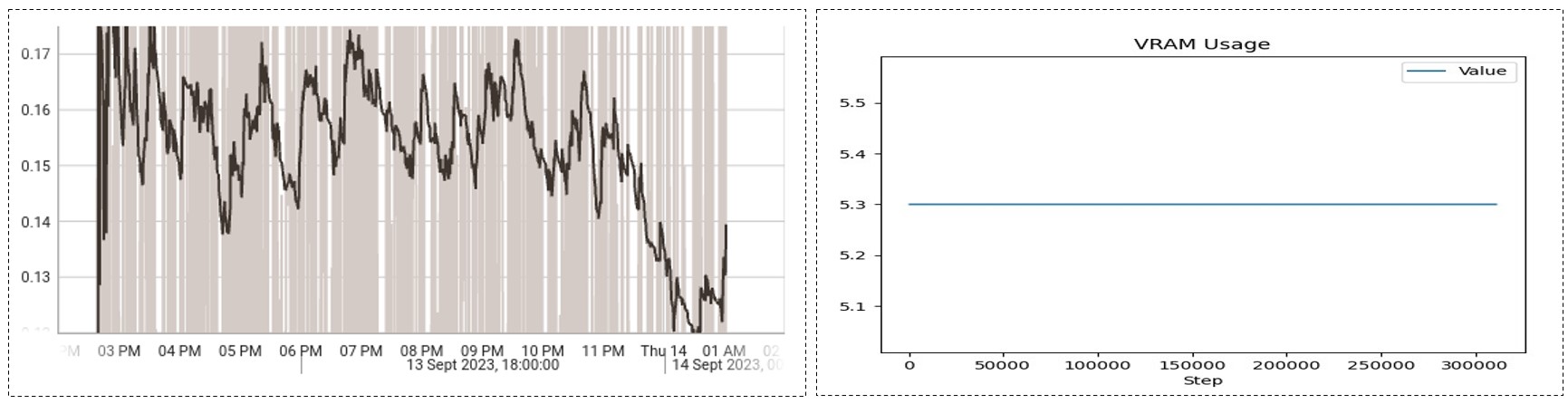}
    \caption{Loss graph of tuned Derm-T2IM with final loss value of 0.1394 and overall training time of 10.4 hours. The second graph shows the average amount of video RAM i.e., 5.3 GB required during the whole training process.}
    \label{fig_3}
\end{figure}
As it can be observed from Table~\ref{tab1} that we have selected Discrete Denoising Scheduler (DDS) as the noise scheduler. The noise schedule is a critical component of diffusion models as it simulates the gradual diffusion process from noisy to clean data. It is important to select an effective noise scheduler to ensure that the model converges and generates realistic data samples. The Discrete Denoising Scheduler (DDS) is a scheduling strategy used in denoising diffusion probabilistic models (DDPMs)~\cite{ho2020denoising} to determine how the noise level should evolve during the denoising process. DDS provides a discrete set of noise levels at each step and helps control the level of noise introduced during the training process. The DDS representation as shown in equation 3 is a discrete formulation used to schedule the noise levels at each training step. This equation defines how the observation at the current time step ($x_t$) is obtained from the previous observation ($x_{t-1}$) by adding noise ($\sigma_t \ast z_t$). The noise level $\sigma_t$ is a key component of the DDS and is typically scheduled to decrease over time during training.

$$
x_t = x_{t-1} + (\sigma_t \ast z_t) \eqno{(3)}
$$

Where $x_t$ is the noisy observation at time step $t$. $x_{t-1}$ is the observation at the previous time step ($t-1$). ${\sigma}_t$ is the noise level or standard deviation at time step $t$. $z_t$ is a sample from the standard normal distribution (mean 0, standard deviation 1). It is designed to be computationally efficient while achieving high-quality sample generation.

As the optimizer, we have selected ‘AdamW’~\cite{loshchilov2017decoupled} with ‘Constant with Warmup’ as the learning scheduler. "Constant with Warmup" is a commonly used learning rate scheduling technique in various deep learning applications. It helps strike a balance between stability and speed in training/ tuning deep learning models. Finally, FP16 model precision is selected. FP16 is a valuable tool in the optimization of deep learning and high-performance computing workloads. When used judiciously and with consideration of its limitations, it can significantly improve the efficiency and speed of computations, making it particularly relevant for large-scale machine learning based image rendering operations and scientific simulations.

\subsection{Image Inference Samplers} \label{sampler}
Once the model is trained using few shot data learning and optimal selection of training parameters, the sixth step as shown in Fig.~\ref{framework} includes rendering new synthetic dermatoscopic data using various image sampling methods. In this research study we have used three different text-to-image sampling methods which incorporate Euler, Euler a and PLMS methods~\cite{Samplers}. Euler sampling stands as the simplest possible sampling method, sharing mathematical equivalence with Euler's method for solving ordinary differential equations. It operates entirely deterministically, devoid of any introduction of random noise during the sampling process. Euler ancestral (Euler a) sampler is like Euler’s sampler. But at each step, it subtracts more noise than it should and adds some random noise back to match the noise schedule. The denoised image depends on the specific noise added in the previous steps. So, it is an ancestral sampler, in the sense that the path the image denoises depends on the specific random noises added in each step. Further we have used a more recent sampling technique known as Pseudo Numerical Methods for Diffusion Models (PLMS) on Manifolds are computational techniques designed to approximate and simulate diffusion processes on mathematical manifolds. These methods are particularly useful for modeling complex data distributions and conducting probabilistic modeling and generative tasks on manifolds. They allow for the estimation of data distribution dynamics and the generation of realistic data samples by discretizing the diffusion process into numerical steps. The term "pseudo" indicates that these methods approximate the behaviour of diffusion processes, typically using numerical integration or sampling techniques.

\section{Experimental Results} \label{experiment}
All our experiments were conducted by integrating dreambooth extension with Stable Diffusion Web User Interface which is based on the Gradio python library for developing customized dermatoscopic text-to-image diffusion models. The complete environment was set up on a workstation machine equipped with the RTX3090Ti graphics card with 24GB of dedicated graphics memory.

\subsection{Fine Tuning Results}

\begin{figure}[thpb]
    \centering
    \includegraphics[width=0.8\linewidth]{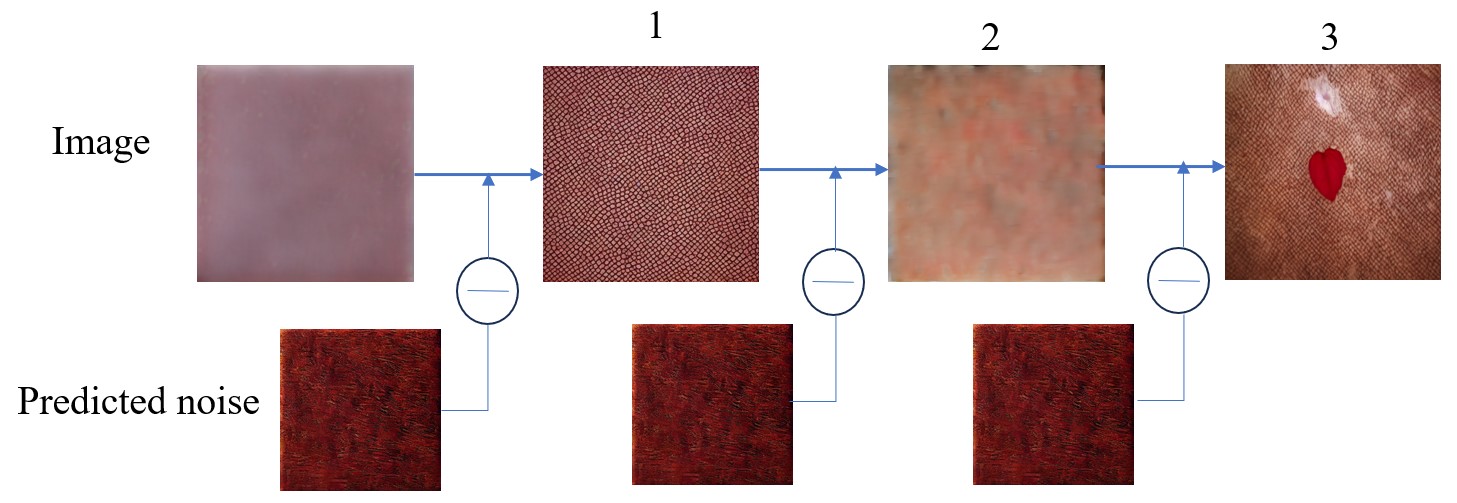}
    \caption{Learning samples with predicted random nosie while performing the fine-tuning process of Derm-T2IM: First Sample extracted at 12990 steps and last sample extracted at 311760 steps.}
    \label{fig_4}
\end{figure}

Initially, we used various training parameters as the ablation study to analyze the model tuning results. Finally, we shortlisted the set of training parameters as mentioned in Table~\ref{tab2} for performing optimal convergence and model generalization on our custom skin lesion data. The model was fine-tuned with a batch size of 2 for 120 epochs. We used a high learning rate of 2e-6 and a low learning rate of 1e-6 and further no prior preservation was used. It is important to discuss that we have used 8-bit Adam, along with fp16 gradient accumulation to reduce memory requirements and run these experiments on GPUs with 24 GB of memory. The overall training process was completed in 10.4 hours. Fig.~\ref{fig_3} shows the loss graph of the tuned model extracted via the Tensorboard visualization tool and the average VRAM required during the fine-tuning process. Whereas Fig.~\ref{fig_4} shows the learning samples during the model fine-tuning process at various steps. Since in this research work, we are more interested in employing a pre-trained text-to-image diffusion model for validating its effectiveness in medical imaging application we have used frozen text encoder resulting in small, personalized token embeddings.

\subsection{Derm-T2IM Rendering Results using different sampling methods}

\begin{figure*}[thpb]
    \centering
    \includegraphics[width=0.9\linewidth]{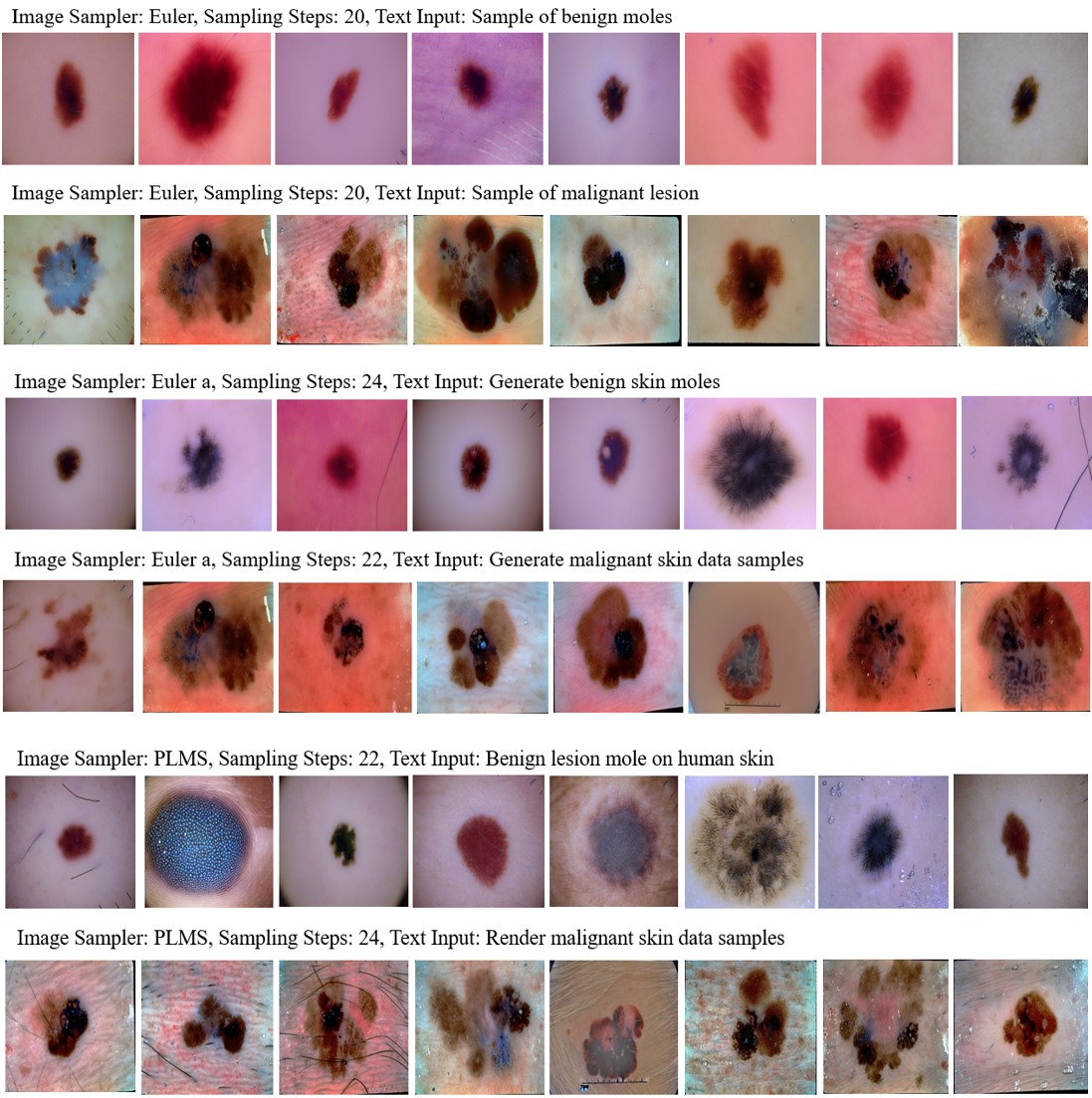}
    \caption{Rendered outputs showing newly generated benign and malignant skin mole data using Derm-T2IM. The first, third and fifth rows show the results of benign data generated via Euler, Euler a, and PLMS sampling method whereas second, fourth and six rows show the malignant lesion inference results generated via Euler, Euler a, and PLMS samplers.}
    \label{fig_5}
\end{figure*}

The fine-tuned Derm-T2IM model is subsequently employed to generate novel synthetic skin lesion data. This is achieved by using three different image sampling methods as discussed in Section~\ref{sampler}. Fig.~\ref{fig_5} shows the results along with user text prompts.

As it can be observed from Fig.~\ref{fig_5} that quality of rendered data improves by increasing by sampling steps. Therefore, we have mostly used sampling steps between the scale of 22-26 for generating large scale skin lesion data. The overall dataset was generated with image resolution of $512 \times 512$ and stored in PNG format. PNG is lossless compression and thus retains all the data in the file during the compression process, which is important when we have to resize the data for further use for downstream machine learning tasks. Furthermore, in the image rendering process, the classifier-free guidance scale (CFG) scale was configured to a value of 7. Typically, this setting determines the degree to which the generated image aligns with the user's input prompt. Higher values tend to maintain the image's proximity to the user's prompt.

\subsection{Smart Image Transformation using Diversified Text Embeddings}

\begin{figure}[thpb]
    \centering
    \includegraphics[width=0.9\linewidth]{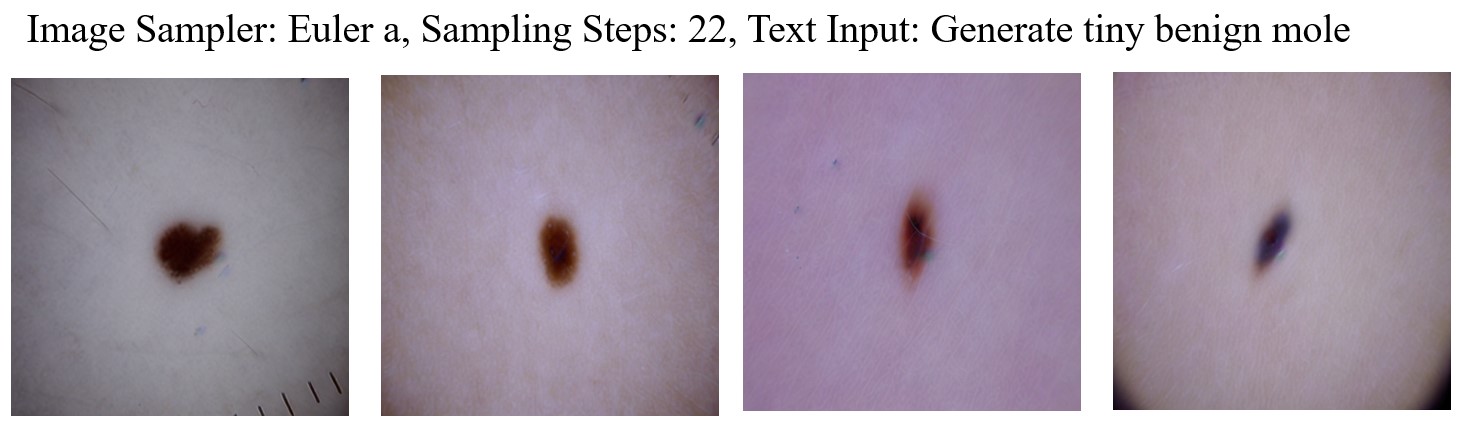}\\(a)\\
    \includegraphics[width=0.9\linewidth]{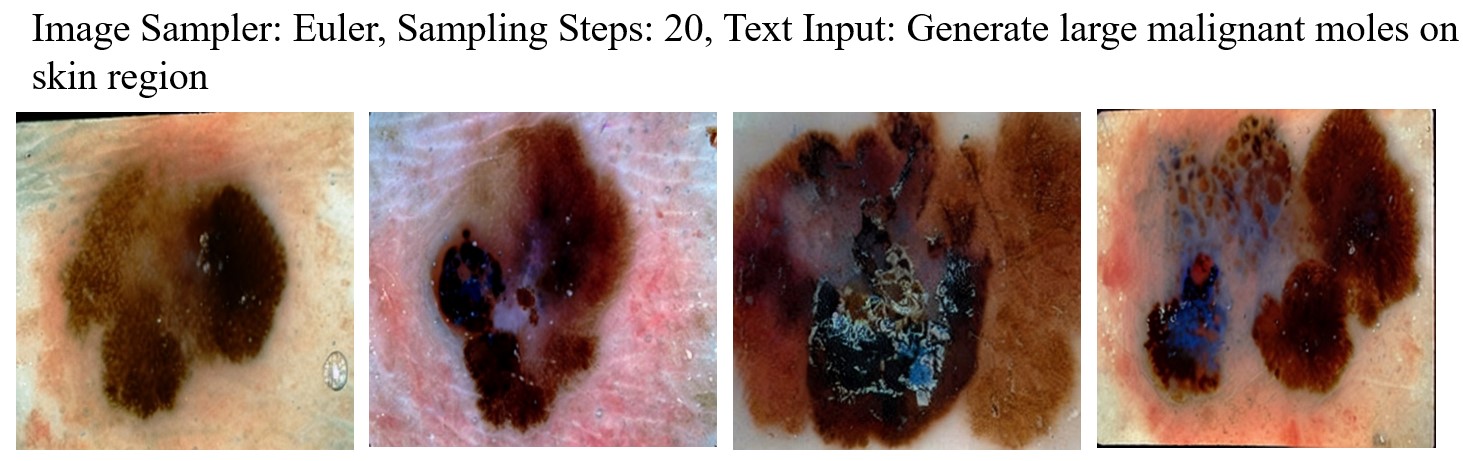}\\(b)\\
    \includegraphics[width=0.9\linewidth]{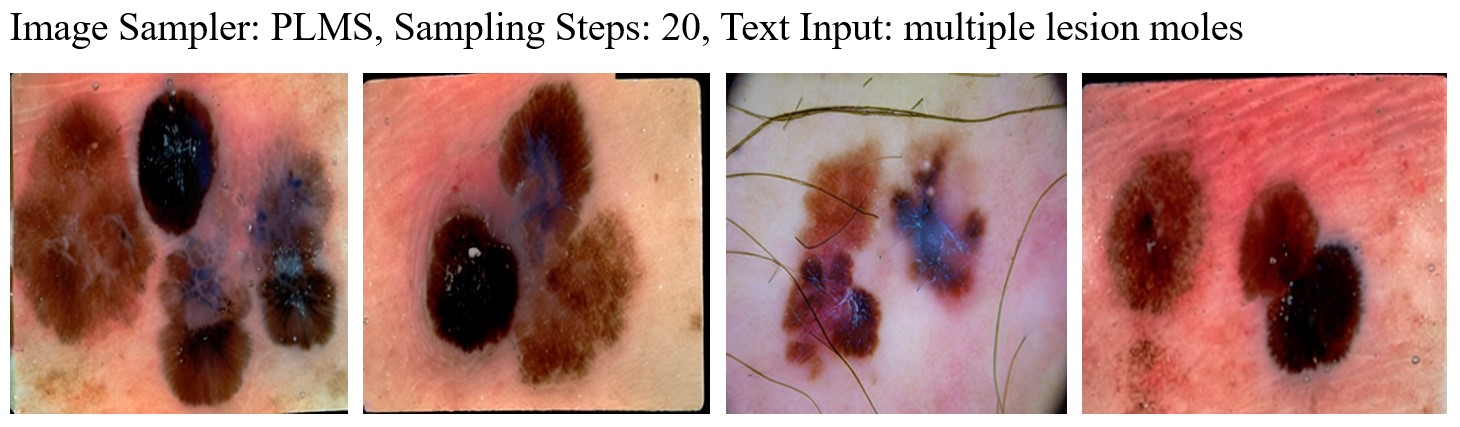}\\(c)\\
    \caption{Derm-T2IM smart transformation outputs: a) shows the tiny benign mole results using Euler a method, b) shows the sample images of large malignant mole, c) shows the transformation results of multiple lesion moles.}
    \label{fig_6}
\end{figure}

To further validate the potential of optimally tuned Derm-T2IM architecture we have produced pigmented lesion data with advanced transformation based on user text inputs/ prompts. These smart data transformations are produced by employing three different sampling methods as discussed in Section~\ref{sampler} and varying sampling steps. The purpose is to generate data that can help the research community and medical professionals to better understand and classify the moles under peculiar circumstances. These transformations include manipulating the size of moles, rendering multiple moles, generating dermatoscopic data embedded on different skin colour. The detailed results are demonstrated in Fig.~\ref{fig_6}. 

It can be observed from Fig.~\ref{fig_6} that a tuned model can be efficiently used for producing a variety of data without undergoing any further specific training/ tuning process. For now, we have tested the Derm-T2IM with these types of advanced augmentations however further diversity can also be rendered depending on specific use cases.

\begin{figure}[thpb]
    \centering
    \includegraphics[width=0.9\linewidth]{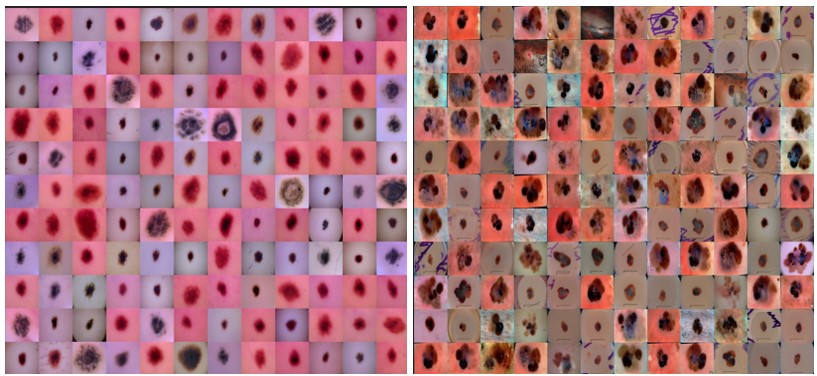}
    \caption{Large scale novel synthetic skin lesion dataset generated using Derm-T2IM Stable Diffusion, a) $11 \times 12$ image grid showing the sample of benign lesion data generated using euler, euler a and PLMS sampling methods, b) show the sample of malignant lesion data generated using Euler, Euler a and PLMS sampling methods.}
    \label{fig_7}
\end{figure}

\subsection{Novel Large Scale Skin Lesion Data and Open-Sourced Derm-T2IM Stable Diffusion Architecture}

Through this work we have open-sourced a large scale synthetic dermatoscopic data by employing the fine-tuned stable diffusion model (Derm-T2IM). This is achieved by storing the trained model using a simpler yet faster safetensors format. In the second phase the Derm-T2IM safetensor file is uploaded via text to image inference engine for rendering novel synthetic skin lesion dataset. The overall dataset consists of binary classes i.e., ‘benign’ and ‘malignant’ moles which is generated by providing the different user prompts as shown in Fig.~\ref{fig_7}. Further  Fig.~\ref{fig_7} shows the image grid of data samples collected from novel synthetic skin lesion data whereas Table~\ref{tab3} shows the complete dataset attributes. 

\begin{table}[!tbp]
\centering
\caption{Datasets details}
\label{tab3}
\resizebox{\linewidth}{!}{
\begin{tabular}{|c|c|c|c|c|}
\hline
\begin{tabular}[c]{@{}c@{}}Data \\ Class\end{tabular} &
  \begin{tabular}[c]{@{}c@{}}No of \\ Samples\end{tabular} &
  \begin{tabular}[c]{@{}c@{}}Sampling \\ Methods\end{tabular} &
  \begin{tabular}[c]{@{}c@{}}Sampling \\ steps\end{tabular} &
  \begin{tabular}[c]{@{}c@{}}Image Resolution\\ and Image\\ compression\end{tabular} \\ \hline
Benign &
  3k &
  \multirow{2}{*}{\begin{tabular}[c]{@{}c@{}}Euler, \\ Euler a,  PLMS\end{tabular}} &
  \multirow{2}{*}{20-26} &
  \multirow{2}{*}{\begin{tabular}[c]{@{}c@{}}$512 \times 512$,\\ PNG\end{tabular}} \\ \cline{1-2}
Malignant &
  3k &
   &
   &
   \\ \hline
\end{tabular}%
}
\end{table}

It's imperative to note that during the generation and storage of a large-scale synthetic skin lesion dataset, we carefully excluded a set of images containing noisy artifacts as demonstrated in Figure 8. This process resulted in a pristine and high-quality dataset. The complete Synthetic Derm-T2IM Dataset is available via our GitHub repository: 

\begin{figure}[thpb]
    \centering
    \includegraphics[width=0.9\linewidth]{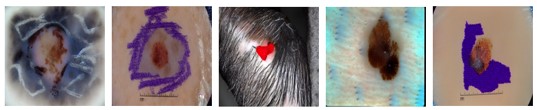}
    \caption{Samples of discarded data with noisy artifacts.}
    \label{fig_8}
\end{figure}

\section{Synthetic Data Validation}

\begin{figure}[thpb]
    \centering
    \includegraphics[width=0.9\linewidth]{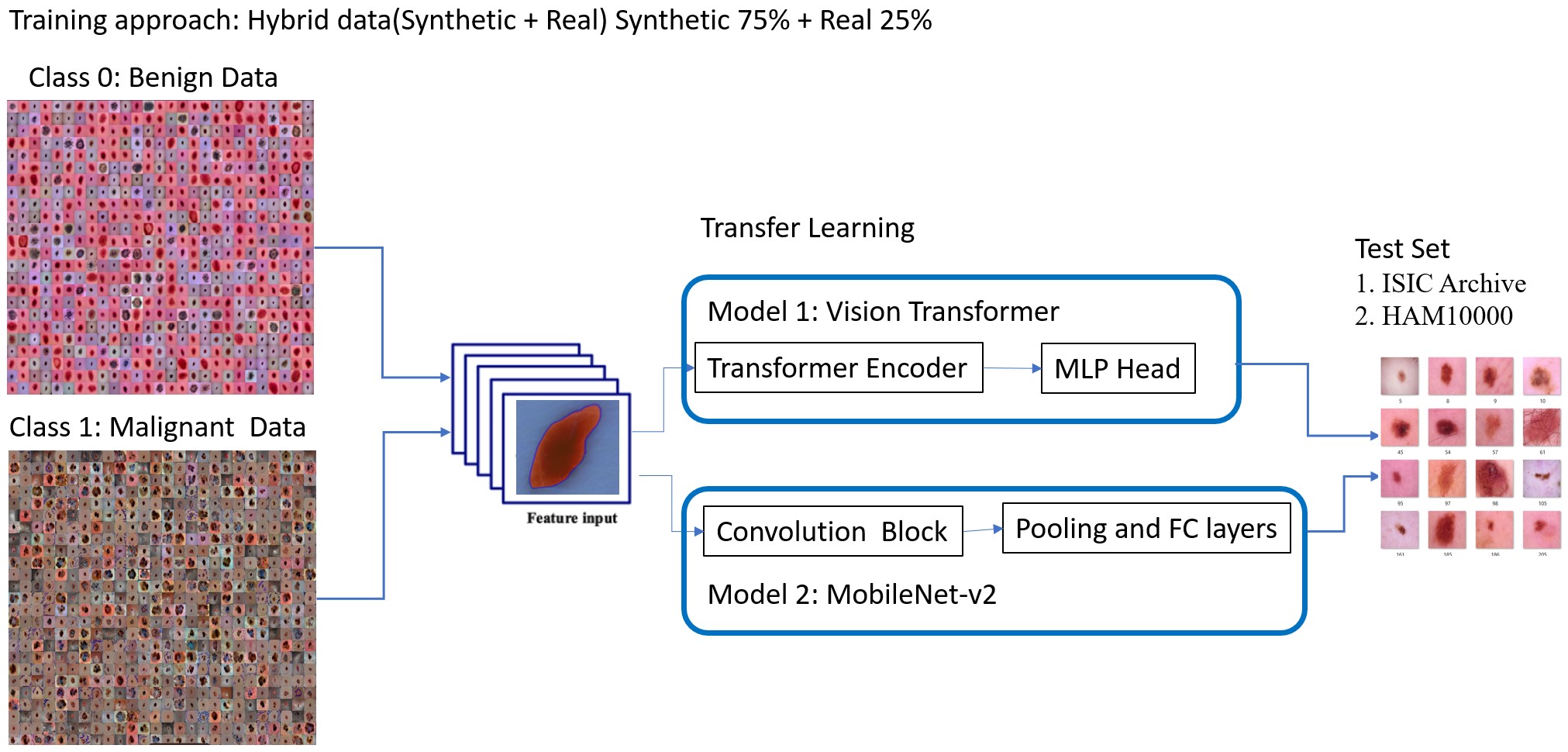}
    \caption{Derm-T2IM synthetic data validation: adapted hybrid training data approach for better fine-tuning of MobileNet-V2 and Vision Transformers Models.}
    \label{fig_9}
\end{figure}

In the second phase of this research study, we aim to assess the quality and utility of the data generated via Derm-T2IM. We do this by training and validating two state-of-the-art skin lesion classifiers, which incorporate Vision Transformer and MobileNet V2 CNN. Vision Transformers (ViTs) have proven to be highly effective for medical image classification tasks. Their ability to capture complex patterns and features within medical images has made them a valuable tool in healthcare and medical research application. In this study we have used the ‘google/vit-base-patch16-224-in21k’ model developed by Google. It's a base-sized model that takes $224 \times 224$ pixel images divided into $16 \times 16$ pixel patches as input and was pre-trained on a dataset with 21,000 classes~\cite{dosovitskiy2020image}. Similarly, MobileNetV2 is a convolutional neural network architecture designed for efficient on-device image classification and computer vision tasks. It introduces the concept of inverted residuals, which involves the use of lightweight depth-wise separable convolutions with linear bottlenecks. This design enhances the efficiency of the network. Further the linear bottlenecks help to maintain low-dimensional embeddings between layers, allowing for better information flow and reducing the risk of information loss. Such pre-trained models can be fine-tuned via transfer learning for various medical imaging applications and as robust feature extractors for image analysis. Fig.~\ref{fig_9} shows the adapted training and validation methodology for both the skin lesion classifiers. 

It can be observed from Fig.~\ref{fig_9} that we have used hybrid data training approach for the purpose of transfer learning and further cross validating the tuned networks on a set of testing data collected from real-world skin cancer datasets. Transfer learning is a powerful technique in deep learning where pre-trained models, initially trained on large-scale datasets, are fine-tuned for new tasks/ domains~\cite{9134370} without the requirement of training the complete network from scratch. When it comes to computer vision, particularly image classification tasks, transfer learning can be applied to both Vision Transformers (ViTs) and Convolutional Neural Networks (CNNs). Hybrid training data comprises a larger division ($\approx 75\%$) of synthetic data with small of portion of real data samples ($\approx 25\%$) to incorporate data diversity, avoid model over fitting, and enlarge the overall training data size. The performance of tuned binary classifier models trained on hybrid data is compared with baseline models which are fine-tuned using only real dermoscopic data samples. For comparison analysis we have perform the cross-validation test on a set of unseen images collected from two different datasets which includes ISIC archive\footnote{https://www.isic-archive.com/ \label{isic}} and HAM10000~\cite{DBW86T_2018} dataset. The below graphs show the accuracy comparison analysis on validation set of training data of both the tuned networks.

\begin{figure}[thpb]
    \centering
    \includegraphics[width=0.9\linewidth]{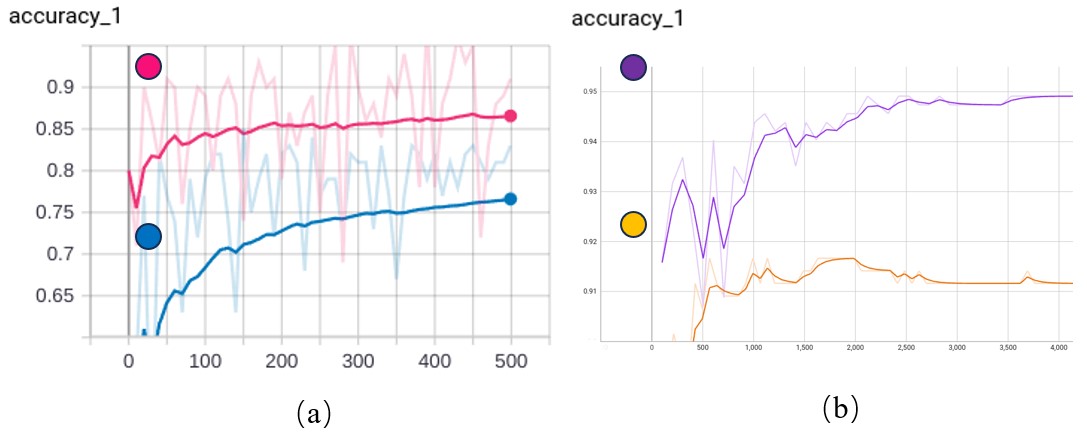}
    \caption{Accuracy comparison computed on the validation set of training data. a) MobileNet: The pink graph shows the model tuned on hybrid data with validation accuracy of 86.57\% whereas the blue graph shows the baseline model trained on only real-world skin data with validation accuracy of 76.61\%. b) ViT: The purple graph shows the model tuned on hybrid data with validation accuracy of 94.74\% whereas the orange graph shows the baseline model trained on only real-world skin data with validation accuracy of 91.16\%}
    \label{fig_10}
\end{figure}

It can be observed from Fig.~\ref{fig_10} that hybrid data approach which comprises of blended mixture of real and synthetic data samples helps in robust tuning of both CNN and vision transformer models.  Further we have used a unseen test dataset acquired from two different real-world skin cancer datasets to perform the rigorous cross validation of both the networks as shown in Table~\ref{tab4}  The results demonstrates that models trained on hybrid data outperforms the baseline models by achieving better testing accuracy with nearly an improvement of 2.46\% in case of mobilenet and 2.29\% in case of ViT model.

\begin{table}[!tbp]
\centering
\caption{Test Accuracy}
\label{tab4}
\resizebox{\linewidth}{!}{
\begin{tabular}{|ll|l|l|}
\hline
\multicolumn{2}{|l|}{} & \begin{tabular}[c]{@{}l@{}}Dataset-1\\ ISIC Archive\end{tabular} & \begin{tabular}[c]{@{}l@{}}Dataset-2\\ HAM10000\end{tabular} \\ \hline
\multicolumn{1}{|l|}{\multirow{2}{*}{\begin{tabular}[c]{@{}l@{}}Baseline Model\\ Accuracy \% \end{tabular}}} & MobileNet & 78.89 & 70.33 \\ \cline{2-4} 
\multicolumn{1}{|l|}{}                                                                                    & ViT       & 90.74 & 81.50 \\ \hline
\multicolumn{1}{|l|}{\multirow{2}{*}{\begin{tabular}[c]{@{}l@{}}Hybrid Model\\ Accuracy \% \end{tabular}}}  & MobileNet & 81.35 & 73.58 \\ \cline{2-4} 
\multicolumn{1}{|l|}{}                                                                                    & ViT       & 93.03 & 85.02 \\ \hline
\end{tabular}%
}
\end{table}

\begin{figure}[thpb]
    \centering
    \includegraphics[width=0.9\linewidth]{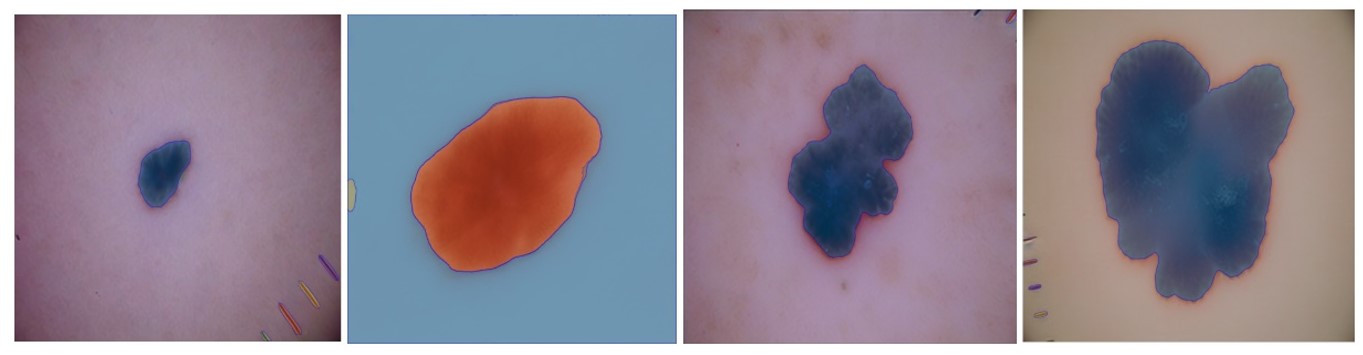}
    \caption{Mole segmentation results of four different cases using zero shot Segment Anything model. The first two images show precise mole boundary and region of interest segmentation on benign data whereas the third and fourth shows the segmentation results on malignant data.}
    \label{fig_11}
\end{figure}

\begin{figure}[thpb]
    \centering
    \includegraphics[width=0.9\linewidth]{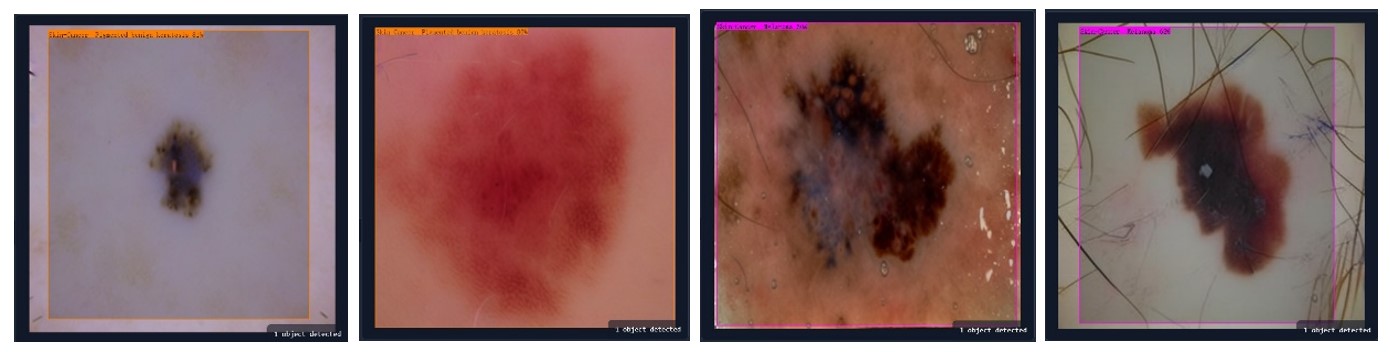}
    \caption{Mole detection and classification results using SDD Yolo-v8 tuned on ISIC archive datasets. The first two images are detected and classified as benign keratosis with 75\% and 86\% confidence scores whereas the third and four cases are classified as melanoma with confidence scores of 60\% and 63\%.}
    \label{fig_12}
\end{figure}

In addition, to this we have performed two other validation tests. The first test includes precise mole segmentation without the need for additional training. For this purpose, we have incorporated zero shot Segment Anything model developed by Meta AI\footnote{https://segment-anything.com/demo \label{mateai}} to perform mole segmentation on Derm2-T2IM dataset samples. The test was carried out on more than 100 different cases from both the classes. The segmentation results demonstrated in Fig.~\ref{fig_11} shows the robustness of pretrained model on novel systematic synthetic datasets. In the second phase, we have performed detection as well as classification test on Derm-T2IM dataset by incorporating pretrained SSD YOLO-V8\footnote{https://github.com/ultralytics/ultralytics \label{yolo8}} model trained on real world ISIC archive dataset\footnote{https://universe.roboflow.com/ai-eds7n/skin-cancer-recogniser \label{pretrainyolo}}. In this case the cross-validation test was performed on 200 random samples selected from Derm-T2IM dataset. The model achieves the overall test accuracy of 72\%. The inferenced result on four different cases shortlisted from synthetic dataset is depicted in Fig.~\ref{fig_12}. 

\section{Conclusion and Future Work}

In this work, we have proposed an efficient text-to-image stable diffusion model based on few-shot learning with a small number of dermoscopic medical data representations for rendering high-quality synthetic skin lesion data. The generated data showcases the superior generative abilities of probabilistic diffusion models in creating diversified skin lesion images of different categories. Further, the efficacy of the generated data is validated through the integration of a hybrid training data approach, involving a blend of small amount of real and synthetic data. This approach enhances the robustness of training Convolutional Neural Network (CNN) and Vision Transformer models using the majority of synthetic data blended with real data. The results demonstrate that utilizing hybrid data contributes to achieving more precise validation and test outcomes on unseen real test data sourced from various publicly available skin cancer datasets. As conclusion the synthetic dermoscopic images are a valuable resource for dermatologists, researchers, and machine learning practitioners. They contribute to the development of more accurate diagnostic tools, better training for healthcare professionals, and the improvement of skin lesion analysis algorithms, ultimately enhancing the early detection and management of skin conditions.

As the possible future research directions, the existing system can be retrained to expand its generating capabilities thus rendering more  heterogeneous data such as specific type of malignant neoplasms, other type of skin diseases such that acne cystica, bullous pemphigoid, and generating data of various degree of burn/ scald cases which involve the entire epidermis, resulting in erythema and blistering. Moreover, we can explore to develop methods to allow users more control over the generation process, enabling interactive and user-guided content synthesis.

\bibliographystyle{unsrt}
\bibliography{main.bib}

\begin{thebibliography}{10}

\bibitem{SkinCancerStatistics}
{American Academy of Dermatology}.
\newblock Skin cancer statistics, 2023.
\newblock \url{https://www.aad.org/media/stats-skin-cancer}, Last accessed on 27-Nov-2023.

\bibitem{Cancerstatistics2023}
Rebecca~L. Siegel, Kimberly~D. Miller, Nikita~Sandeep Wagle, and Ahmedin Jemal.
\newblock Cancer statistics, 2023.
\newblock {\em CA: A Cancer Journal for Clinicians}, 73(1):17--48, 2023.

\bibitem{Burden}
Martijn Meijs, Astrid Herrera, Alvaro Acosta, and Esther Vries.
\newblock Burden of skin cancer in colombia.
\newblock {\em International Journal of Dermatology}, 61, 02 2022.

\bibitem{rashid2022skin}
Javed Rashid, Maryam Ishfaq, Ghulam Ali, Muhammad~R Saeed, Mubasher Hussain, Tamim Alkhalifah, Fahad Alturise, and Noor Samand.
\newblock Skin cancer disease detection using transfer learning technique.
\newblock {\em Applied Sciences}, 12(11):5714, 2022.

\bibitem{pathania2022non}
Yashdeep~Singh Pathania, Zoe Apalla, Gabriel Salerni, Anant Patil, Stephan Grabbe, and Mohamad Goldust.
\newblock Non-invasive diagnostic techniques in pigmentary skin disorders and skin cancer.
\newblock {\em Journal of cosmetic dermatology}, 21(2):444--450, 2022.

\bibitem{ricotti2009malignant}
Carlos Ricotti, Navid Bouzari, Amar Agadi, and Clay~J Cockerell.
\newblock Malignant skin neoplasms.
\newblock {\em Medical Clinics}, 93(6):1241--1264, 2009.

\bibitem{carlini2023extracting}
Nicolas Carlini, Jamie Hayes, Milad Nasr, Matthew Jagielski, Vikash Sehwag, Florian Tramer, Borja Balle, Daphne Ippolito, and Eric Wallace.
\newblock Extracting training data from diffusion models.
\newblock In {\em 32nd USENIX Security Symposium (USENIX Security 23)}, pages 5253--5270, 2023.

\bibitem{saharia2022photorealistic}
Chitwan Saharia, William Chan, Saurabh Saxena, Lala Li, Jay Whang, Emily~L Denton, Kamyar Ghasemipour, Raphael Gontijo~Lopes, Burcu Karagol~Ayan, Tim Salimans, et~al.
\newblock Photorealistic text-to-image diffusion models with deep language understanding.
\newblock {\em Proc. Int. Conf. Neural Inf. Process. Syst.}, 35:36479--36494, 2022.

\bibitem{wang2020generalizing}
Yaqing Wang, Quanming Yao, James~T Kwok, and Lionel~M Ni.
\newblock Generalizing from a few examples: A survey on few-shot learning.
\newblock {\em ACM computing surveys (csur)}, 53(3):1--34, 2020.

\bibitem{rombach2021highresolution}
Robin Rombach, Andreas Blattmann, Dominik Lorenz, Patrick Esser, and Björn Ommer.
\newblock High-resolution image synthesis with latent diffusion models, 2021.

\bibitem{ruiz2023dreambooth}
Nataniel Ruiz, Yuanzhen Li, Varun Jampani, Yael Pritch, Michael Rubinstein, and Kfir Aberman.
\newblock Dreambooth: Fine tuning text-to-image diffusion models for subject-driven generation.
\newblock In {\em Proceedings of the IEEE/CVF Conference on Computer Vision and Pattern Recognition}, pages 22500--22510, 2023.

\bibitem{dosovitskiy2020image}
Alexey Dosovitskiy, Lucas Beyer, Alexander Kolesnikov, Dirk Weissenborn, Xiaohua Zhai, Thomas Unterthiner, Mostafa Dehghani, Matthias Minderer, Georg Heigold, Sylvain Gelly, et~al.
\newblock An image is worth 16x16 words: Transformers for image recognition at scale.
\newblock {\em arXiv preprint arXiv:2010.11929}, 2020.

\bibitem{howard2017mobilenets}
Andrew~G Howard, Menglong Zhu, Bo~Chen, Dmitry Kalenichenko, Weijun Wang, Tobias Weyand, Marco Andreetto, and Hartwig Adam.
\newblock Mobilenets: Efficient convolutional neural networks for mobile vision applications.
\newblock {\em arXiv preprint arXiv:1704.04861}, 2017.

\bibitem{jones2022artificial}
OT~Jones, RN~Matin, M~van~der Schaar, K~Prathivadi Bhayankaram, CKI Ranmuthu, MS~Islam, D~Behiyat, R~Boscott, N~Calanzani, Jon Emery, et~al.
\newblock Artificial intelligence and machine learning algorithms for early detection of skin cancer in community and primary care settings: a systematic review.
\newblock {\em The Lancet Digital Health}, 4(6):e466--e476, 2022.

\bibitem{farooq2016automatic}
Muhammad~Ali Farooq, Muhammad Aatif~Mobeen Azhar, and Rana~Hammad Raza.
\newblock Automatic lesion detection system (alds) for skin cancer classification using svm and neural classifiers.
\newblock In {\em 2016 IEEE 16th International Conference on Bioinformatics and Bioengineering (BIBE)}, pages 301--308. IEEE, 2016.

\bibitem{bhatt2023state}
Harsh Bhatt, Vrunda Shah, Krish Shah, Ruju Shah, and Manan Shah.
\newblock State-of-the-art machine learning techniques for melanoma skin cancer detection and classification: a comprehensive review.
\newblock {\em Intelligent Medicine}, 3(03):180--190, 2023.

\bibitem{gairola2022exploring}
Ajay~Krishan Gairola, Vidit Kumar, and Ashok~Kumar Sahoo.
\newblock Exploring the strengths of pre-trained cnn models with machine learning techniques for skin cancer diagnosis.
\newblock In {\em 2022 IEEE 2nd Mysore Sub Section International Conference (MysuruCon)}, pages 1--6. IEEE, 2022.

\bibitem{Melanoma}
{International Skin Imaging Collaboration: Melanoma Project}, 2023.
\newblock \url{https://isic-archive.com}, Last accessed on 27-Nov-2023.

\bibitem{mendoncca2013ph}
Teresa Mendon{\c{c}}a, Pedro~M Ferreira, Jorge~S Marques, Andr{\'e}~RS Marcal, and Jorge Rozeira.
\newblock Ph 2-a dermoscopic image database for research and benchmarking.
\newblock In {\em 2013 35th annual international conference of the IEEE engineering in medicine and biology society (EMBC)}, pages 5437--5440. IEEE, 2013.

\bibitem{tschandl2018ham10000}
Philipp Tschandl, Cliff Rosendahl, and Harald Kittler.
\newblock The ham10000 dataset, a large collection of multi-source dermatoscopic images of common pigmented skin lesions.
\newblock {\em Scientific data}, 5(1):1--9, 2018.

\bibitem{inbook}
Lucia Ballerini, Robert Fisher, Ben Aldridge, and Jonathan Rees.
\newblock {\em A Color and Texture Based Hierarchical K-NN Approach to the Classification of Non-melanoma Skin Lesions}, volume~6, pages 63--86.
\newblock 01 2013.

\bibitem{Dermatology}
{Dermatology Information System}, 2023.
\newblock \url{https://www.dermis.net/dermisroot/en/home/index.htm}, Last accessed on 27-Nov-2023.

\bibitem{skandarani2023gans}
Youssef Skandarani, Pierre-Marc Jodoin, and Alain Lalande.
\newblock Gans for medical image synthesis: An empirical study.
\newblock {\em Journal of Imaging}, 9(3):69, 2023.

\bibitem{pesteie2019adaptive}
Mehran Pesteie, Purang Abolmaesumi, and Robert~N Rohling.
\newblock Adaptive augmentation of medical data using independently conditional variational auto-encoders.
\newblock {\em IEEE transactions on medical imaging}, 38(12):2807--2820, 2019.

\bibitem{kazerouni2023diffusion}
Amirhossein Kazerouni, Ehsan~Khodapanah Aghdam, Moein Heidari, Reza Azad, Mohsen Fayyaz, Ilker Hacihaliloglu, and Dorit Merhof.
\newblock Diffusion models in medical imaging: A comprehensive survey.
\newblock {\em Medical Image Analysis}, page 102846, 2023.

\bibitem{shin2018medical}
Hoo-Chang Shin, Neil~A Tenenholtz, Jameson~K Rogers, Christopher~G Schwarz, Matthew~L Senjem, Jeffrey~L Gunter, Katherine~P Andriole, and Mark Michalski.
\newblock Medical image synthesis for data augmentation and anonymization using generative adversarial networks.
\newblock In {\em Simulation and Synthesis in Medical Imaging: Third International Workshop, SASHIMI 2018, Held in Conjunction with MICCAI 2018, Granada, Spain, September 16, 2018, Proceedings 3}, pages 1--11. Springer, 2018.

\bibitem{nie2018medical}
Dong Nie, Roger Trullo, Jun Lian, Li~Wang, Caroline Petitjean, Su~Ruan, Qian Wang, and Dinggang Shen.
\newblock Medical image synthesis with deep convolutional adversarial networks.
\newblock {\em IEEE Transactions on Biomedical Engineering}, 65(12):2720--2730, 2018.

\bibitem{akrout2023diffusion}
Mohamed Akrout, B{\'a}lint Gyepesi, P{\'e}ter Holl{\'o}, Adrienn Po{\'o}r, Bl{\'a}ga Kincs{\H{o}}, Stephen Solis, Katrina Cirone, Jeremy Kawahara, Dekker Slade, Latif Abid, et~al.
\newblock Diffusion-based data augmentation for skin disease classification: Impact across original medical datasets to fully synthetic images.
\newblock {\em arXiv preprint arXiv:2301.04802}, 2023.

\bibitem{dalmaz2022resvit}
Onat Dalmaz, Mahmut Yurt, and Tolga {\c{C}}ukur.
\newblock Resvit: Residual vision transformers for multimodal medical image synthesis.
\newblock {\em IEEE Transactions on Medical Imaging}, 41(10):2598--2614, 2022.

\bibitem{liang2022sketch}
Jiamin Liang, Xin Yang, Yuhao Huang, Haoming Li, Shuangchi He, Xindi Hu, Zejian Chen, Wufeng Xue, Jun Cheng, and Dong Ni.
\newblock Sketch guided and progressive growing gan for realistic and editable ultrasound image synthesis.
\newblock {\em Medical Image Analysis}, 79:102461, 2022.

\bibitem{li2022neural}
Wei Li, Shiping Wen, Kaibo Shi, Yin Yang, and Tingwen Huang.
\newblock Neural architecture search with a lightweight transformer for text-to-image synthesis.
\newblock {\em IEEE Transactions on Network Science and Engineering}, 9(3):1567--1576, 2022.

\bibitem{de2023medical}
Bram de~Wilde, Anindo Saha, Richard~PG ten Broek, and Henkjan Huisman.
\newblock Medical diffusion on a budget: textual inversion for medical image generation.
\newblock {\em arXiv preprint arXiv:2303.13430}, 2023.

\bibitem{shavlokhova2023finetuning}
Veronika Shavlokhova, Andreas Vollmer, Christos~C Zouboulis, Michael Vollmer, Jakob Wollborn, Gernot Lang, Alexander K{\"u}bler, Stefan Hartmann, Christian Stoll, Elisabeth Roider, et~al.
\newblock Finetuning of glide stable diffusion model for ai-based text-conditional image synthesis of dermoscopic images.
\newblock {\em Frontiers in Medicine}, 10, 2023.

\bibitem{ho2020denoising}
Jonathan Ho, Ajay Jain, and Pieter Abbeel.
\newblock Denoising diffusion probabilistic models.
\newblock {\em Proc. Int. Conf. Neural Inf. Process. Syst.}, 33:6840--6851, 2020.

\bibitem{rombach2022high}
Robin Rombach, Andreas Blattmann, Dominik Lorenz, Patrick Esser, and Bj{\"o}rn Ommer.
\newblock High-resolution image synthesis with latent diffusion models.
\newblock In {\em Proceedings of the IEEE/CVF conference on computer vision and pattern recognition}, pages 10684--10695, 2022.

\bibitem{lee1997dullrazor}
Tim Lee, Vincent Ng, Richard Gallagher, Andrew Coldman, and David McLean.
\newblock Dullrazor{\textregistered}: A software approach to hair removal from images.
\newblock {\em Computers in biology and medicine}, 27(6):533--543, 1997.

\bibitem{hu2022lora}
Edward~J Hu, Yelong Shen, Phillip Wallis, Zeyuan Allen-Zhu, Yuanzhi Li, Shean Wang, Lu~Wang, and Weizhu Chen.
\newblock Lo{RA}: Low-rank adaptation of large language models.
\newblock In {\em Proc. Int. Conf. Learn. Representations}, 2022.

\bibitem{loshchilov2017decoupled}
Ilya Loshchilov and Frank Hutter.
\newblock Decoupled weight decay regularization.
\newblock {\em arXiv preprint arXiv:1711.05101}, 2017.

\bibitem{Samplers}
{Samplers in Stable Diffusion}, 2023.
\newblock \url{https://www.felixsanz.dev/articles/complete-guide-to-samplers-in-stable-diffusion#:~:text=DDPM%20(paper)%20(Denoising%20Diffusion,to%20achieve%20a%20decent%20result}, Last accessed on 27-Nov-2023.

\bibitem{9134370}
Fuzhen Zhuang, Zhiyuan Qi, Keyu Duan, Dongbo Xi, Yongchun Zhu, Hengshu Zhu, Hui Xiong, and Qing He.
\newblock A comprehensive survey on transfer learning.
\newblock {\em Proc. IEEE}, 109(1):43--76, 2021.

\bibitem{DBW86T_2018}
Philipp Tschandl.
\newblock {The HAM10000 dataset, a large collection of multi-source dermatoscopic images of common pigmented skin lesions}, 2018.

\end{thebibliography}

\end{document}